\documentclass[a4paper]{article}
\usepackage{dx23}
\usepackage[T1]{fontenc}
\usepackage{ae,aecompl}
\usepackage{amsfonts}
\usepackage{amsmath}
\usepackage{amsthm}
\usepackage{booktabs}
\usepackage{times}

\usepackage{graphicx}
\usepackage{algpseudocode}
\usepackage{array}
\usepackage{algorithm}

\newtheorem{definition}{Definition}

\newcommand{\nat}{\ensuremath{\mathbf{N} }}

\newcommand{\real}{\ensuremath{\mathbf{R} }}

\usepackage{tikz}
\def\checkmark{\tikz\fill[scale=0.4](0,.35) -- (.25,0) -- (1,.7) -- (.25,.15) -- cycle;} 

\title{A Diagnosis Algorithms for a Rotary Indexing Machine}

\author%
{%
Maria Krantz$^1$ \and Oliver Niggemann$^2$\\
$^1$Helmut-Schmidt-University, Hamburg, Germany\\
e-mail: maria.krantz@hsu-hh.de\\
$^2$Helmut-Schmidt-University, Hamburg, Germany\\
e-mail: oliver.niggemann@hsu-hh.de\\
}

\begin{document}

\twocolumn

\maketitle

\begin{abstract}
Rotary Indexing Machines (RIMs) are widely used in manufacturing due to their ability to perform multiple production steps on a single product without manual repositioning, reducing production time and improving accuracy and consistency.
Despite their advantages, little research has been done on diagnosing faults in RIMs, especially from the perspective of the actual production steps carried out on these machines. Long downtimes due to failures are problematic, especially for smaller companies employing these machines.
To address this gap, we propose a diagnosis algorithm based on the product perspective, which focuses on the product being processed by RIMs. The algorithm traces the steps that a product takes through the machine and is able to diagnose possible causes in case of failure. We also analyze the properties of RIMs and how these influence the diagnosis of faults in these machines.
Our contributions are three-fold. Firstly, we provide an analysis of the properties of RIMs and how they influence the diagnosis of faults in these machines. Secondly, we suggest a diagnosis algorithm based on the product perspective capable of diagnosing faults in such a machine. Finally, we test this algorithm on a model of a rotary indexing machine, demonstrating its effectiveness in identifying faults and their root causes.
\end{abstract}

\section{Introduction}

Cyber-Physical Production Systems (CPPS) combine physical systems, such as machines and production lines, with cyber systems, such as sensors, software, and networks, to create an interconnected production environment \cite{monostori2014cyber}. CPPS offer many benefits, including improved efficiency, flexibility, and quality control. However, as with any complex system, CPPS can experience faults that can have significant consequences on the production process and lead to long downtimes \cite{galaske2016disruption}.

Therefore, diagnosing faults in CPPS is crucial to ensure the smooth operation and maintenance of the production system. Diagnosis refers to the process of identifying the root cause of a problem of fault in a system \cite{grastien2013spectrum}. In modern CPPS, this task is often too complex to be performed by humans. Therefore, computer-based methods are used to identify the root cause of a failure. These methods are often based on methods of Artificial Intelligence (AI) \cite{fernandes2022machine}. 

Model-based diagnosis is a common approach used in CPPS to diagnose faults \cite{de2003fundamentals}. The basic idea is to compare the observed behavior of the system with a model of its expected behavior. The model represents the normal or expected behavior of the system and can be used to detect deviations from this behavior that may indicate a fault in the system.

To implement model-based diagnosis in CPPS, a model of the system is necessary. This model can be based on physical laws and equations or can be a more abstract representation of the system's behavior. The model should capture the essential features of the system and should be able to predict its behavior.

Once the model is developed, it is used to compare the actual behavior of the system with the expected behavior. This is done by collecting sensor data and from the system and comparing it with the predictions of the model. Once a fault is detected, its root cause can be identified \cite{narasimhan2007model}. Model-based diagnosis can detect faults that may not be easily detected by other methods and can therefore provide insights into the root cause of the problem. 
\\\\

Rotary Indexing Machines (RIMs) are a type of machine used in manufacturing processes that automatically indexes (rotates) a product to a new position for the next operation to be performed on it.
The basic principle behind RIMs is the use of a circular indexing table that rotates about a vertical or horizontal axis, stopping at certain fixed positions. At these positions, the tools used to transform the product are installed. The indexing table can hold multiple products, which are then positioned at the tool stations for processing.

One of the advantages of RIMs is their ability to perform multiple production steps on a single product without the need for manual repositioning, handling or tool exchange. This reduces production time and improves accuracy and consistency in the production process. RIMs can also be customized with various tools to fit the specific needs of the production process. As such, RIMs are often custom-made for a specific application.

However, RIMs also have some limitations, such as high initial investment costs and limited flexibility in production setup once the machines is installed. Additionally, the machines may require significant maintenance and downtime for repairs and maintenance.
\\\\
Despite the widespread use of Rotary Indexing Machines in production, relatively little research is done on diagnosis of failures in these machines, especially from a perspective of the actual production steps carried out on these machines. However, especially for smaller companies employing these machines long downtime due to failures are problematic. 

Based on these observations, we formulated the following research questions: 

\noindent \emph{RQ1:} Which properties distinguish RIMs from other production machines and how does this affect the diagnosis algorithm needed for these machines?

\noindent \emph{RQ2:} Can an SMT solver be employed to build a diagnosis algorithm for RIMs?

\noindent \emph{RQ3:} Which types of faults can be diagnosed with such an algorithm?

This paper proposes a diagnosis algorithms for RIMs, which focuses on the product processed by these machines. Based on an SMT solver, this algorithm traces the steps a product takes through the machine and is able to diagnose possible causes in case of failure. 

The contributions of this paper are:
\begin{enumerate}
    \item We provide an analysis of the properties of RIMs and how these influence the diagnosis of faults in these machines. 
    \item We suggest a diagnosis algorithm based on the product perspective capable of diagnosing faults in such a machine.
    \item We test this algorithm on a model of a rotary indexing machine.
\end{enumerate}

As an example, a RIM to be constructed by a German company is analyzed.

The remainder of the paper is structured as follows. Section II will present a short survey of the State of the Art of diagnosis of rotary indexing machines. In Section III the characteristics of RIMs will be analyzed and it will be examined how these can be exploited for diagnosis. Furthermore, this section will introduce the model of the RIM used for development and evaluation of the diagnosis algorithms. Section IV will first formalize the problem and then describe the algorithms, followed by an evaluation in Section V. Section VI will close with a conclusion and a short outlook. 

\section{State of the Art} 
\label{sec:sota}

Some research has been done for diagnosis of rotary indexing machines. Most work however does not focus on the products or the processes taking part on the rotary indexing table, but on the rotational parts of the machine. Most of these approaches are based on Machine Learning (ML) approaches, since they are mostly based on vibration signals. 

The work presented in \cite{saha2022development} discusses an intelligent fault diagnosis technique that was developed to diagnose various faults in a deep groove ball bearing, which is an essential component of a rotating machine. The study used an experimental setup to generate faulty data, including inner race fault, outer race fault, and cage fault, along with the healthy condition. The time waveform of raw vibration data was transformed into a frequency spectrum using the fast Fourier transform (FFT) method and analyzed to detect the defective bearing. The study also applied a machine learning algorithm, the support vector machine (SVM), for fault diagnosis.

Similarly, the work presented in \cite{kolar2020fault} uses  a novel deep-learning-based technique for fault diagnosis in rotary machinery. The proposed technique inputs raw three-axis accelerometer signals as a high-definition 1D image into convolutional neural network (CNN) layers that automatically extract signal features, enabling high classification accuracy. The study achieved effective classification of different rotary machinery states using CNN for classification of raw three-axis accelerometer signals.

Furthermore, \cite{rajabi2022fault} proposes a novel approach for fault diagnosis in rotating equipment using permutation entropy, signal processing, and artificial intelligence. Vibration signals are used to detect the faulty state of the bearing and determine the fault type in two separate steps. Permutation entropy is used for fault detection and wavelet packet transform and envelope analysis are used for fault isolation. The method uses a multi-output adaptive neuro-fuzzy inference system classifier to decide about the faulty bearing’s condition by extracting the proper features of the signals. The approach is evaluated using the Case Western Reserve University dataset and shows improved accuracy in diagnosing faults in rotating equipment compared to existing approaches.

Diagnosis of production systems that are not specifically rotary indexing machines can also be carried out using ML-approaches, but are often based on Model-Based Diagnosis approaches. 
\cite{bunte2019model} suggest a new approach to Model-Based Diagnosis for CPS. The approach uses a learned quantitative model to derive residuals for generating a diagnosis model for root cause identification. This approach has advantages such as easy integration of new machine learning algorithms, seamless integration of qualitative models, and significant speed-up of diagnosis runtime. The paper defines the approach, discusses its advantages and disadvantages, and presents real-world use cases for evaluation.

In \cite{diedrich2019model} a new model-based diagnosis approach for detecting and isolating faults in hybrid systems was presented. The approach involves modelling dynamic system behaviour using state space models and calculating Boolean residuals through an observer-pattern. The observer pattern is implemented using a symbolic system description specified in satisfiability theory modulo linear arithmetic. The residuals are used as fault symptoms, and the minimum cardinality diagnosis is obtained using Reiter's diagnosis lattice. This approach has the advantage of automating the diagnosis process and decoupling modelling and diagnosis. The paper also presents an evaluation of the approach using a four-tank model.

This approach was further developed in \cite{balzereit2020automated}, presenting a novel approach for the automated reconfiguration of CPPS using residual-based fault detection and logical calculi. The approach operates on observed system data and information about the system topology to draw causal coherences, reducing modeling efforts. This automated reconfiguration is needed for autonomous systems, as the software controlling the system is often unable to adapt to unforeseen events and faults. The effectiveness of the approach is evaluated using a simulation of a CPPS, namely a tank-model from process engineering. 

An algorithm that focuses on anomaly detection and diagnosis in a manufacturing process was suggested in \cite{saez2019context}. This paper proposes a hybrid model for cyber-physical manufacturing systems (CPMS) that combines sensor data, context information, and expert knowledge to improve anomaly detection and diagnosis. The model uses a multimodel framework and context-sensitive adaptive threshold limits for anomaly detection, and classification models with expert knowledge for root cause diagnosis. The proposed approach was implemented using IoT to extract data from a computer numerical control machine, and results showed that context-sensitive modeling allowed for combining physics-based and data-driven models to detect anomalies and identify root causes such as worn or broken tools or wrong material.

Specifically for RIMs, research has been done on the human-machine interaction for automation, especially in the case of fault. The specific objective of the work done in \cite{chiranth2020assembly} was to design and develop a low-cost and user-friendly Human Machine Interface (HMI) for an automated wheel assembly unit of the back wheel for a kids' swing car. The wheel assembly process is accomplished using a five-station automated rotary index table. Various electro-mechanical elements like proximity sensors, pneumatic cylinders, solenoid valves, and stepper motors are employed for automating the rotary table. The HMI utilizes a low-cost Arduino controller and a touch screen display for real-time monitoring, however, no anomaly detection or diagnosis algorithm is supplied. 

\section{Characteristics of Rotary Indexing Machines}

\begin{figure}[h!]
  \centering
  \includegraphics[width=.5\textwidth]{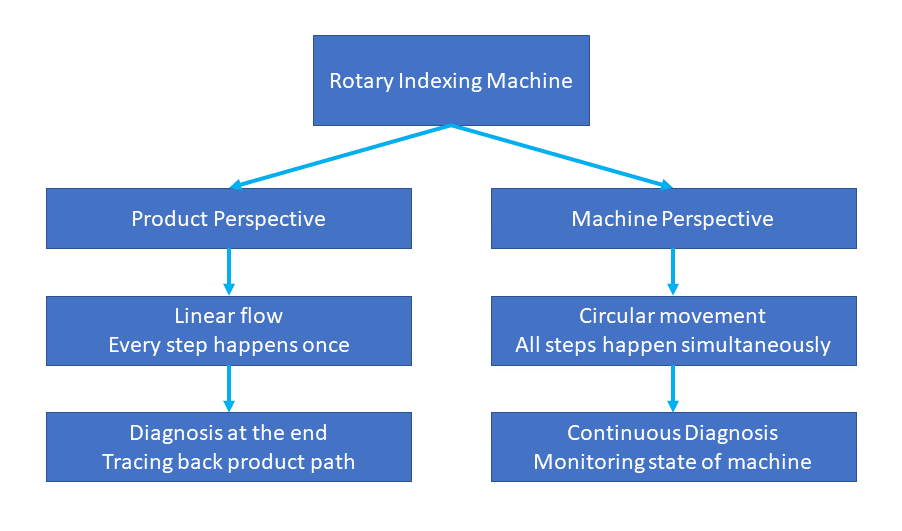}
  \caption{\label{Abb_RIM} There are two different perspectives on diagnosis of Rotary Indexing Machines. One possibility is to monitor the state of the machines, which is often focused on the rotating parts of the machine, as these have the most wear. The other option is to look at the quality of the products processed on the machine. In this case, a damaged product at the end of its processing cycle would indicate a fault in the machine in the preceding steps.}
\end{figure}

As described in Figure 1, there are two possible perspectives when diagnosing RIMs. The first is the perspective of the whole machine, which monitors the state of the moving parts of the machine, often focused on the rotating parts and the bearings. The analysis of these types of failures is interesting, because it is possible to predict the failures of, for example, bearings from vibrational data. However, this is not the focus of this paper. Instead, this work will focus on the second possible perspective, which is that of the product. While a RIM performs a rotating motion, the path of the product through the machine is linear, since the product enters the machine at one station, then goes through each station once and is ejected from the machine one station before the one where it entered. We therefore have a linear product path on a rotating machine. Furthermore, while the machine performs all steps simultaneously on multiple products, a product only sees each station once. 

The machine used as an example here is a RIM with eight stations that assembles a product, performs a quality check and sorts product into OK/Not-OK. The first station is an input station for the first part of the product. This is then followed by several assembly stations, in combination with feeding of further product parts. Afterwards, two stations perform quality control on the product, followed by a station ejecting the OK products and a station ejecting the Not-OK products. Then, the cycle begins anew. 

\subsection{Model of the Machine}

To analyse the RIM and to later test the diagnosis algorithm, two models of the machine were developed. Both models are written in python. Since they are supposed to recreate the data from the actual machine, no effects like vibrations or friction were integrated into the models, because these would also not be reported in the real machine. 

The first model is constructed from the perspective of the product, which moves through each station once. The model simulates one run of a product through the machine. 
The output of the model looks like the following.

\noindent \textit{Thu Apr 27 11:18:58 2023   pneumatic cylinder in position 0} 

The output of the model is the current state of the station the product is at. It consists of a time stamp and the report of the state of the tool currently processing the product. In the real machine, this state would be reported by sensors, e.g. measuring the position of the tools. The model runs through all stations of the machine once. 

The second model simulates the production process of the whole machine. This means that all stations are active simultaneously and several products are processed. The model continuously outputs the state of each station, similar to how the machine would report its internal state during the process cycle. 

\subsection{Diagnosis of Rotary Indexing Machines}

As mentioned above, this work will focus on diagnosis of RIMs from the perspective of the product. While vibrations and damages to rotating parts will impact the quality of the product, many industrially installed RIMs do not have this type of sensor. Instead, sensors will be mostly installed for two main reasons:

\begin{enumerate}
    \item To measure the position of tooling equipment, which is mainly needed as a reporter to the automation software to ensure a tool has reached a designated position before the next step is started.
    \item For quality control purposes. In case the machine features one or more sorting stations, this data is used to sort the product into OK/Not-OK. 
\end{enumerate}

Since this project aimed at developing a diagnosis algorithm for this type of RIMs, the algorithm should only make use of the existing sensors. 

\begin{table}[]
 \label{solutionstable}
 \caption{Characteristics of RIMs which can be employed for diagnosis}
 \centering
\begin{center}
\begin{tabular}{||c | c ||} 
 \hline
 Characteristic of RIM & Use in Diagnosis \\ [0.5ex] 
 \hline\hline
 Rotary Motion & \begin{tabular}[c]{@{}c@{}} Not important for diagnosis \\ from product perspective \end{tabular}  \\ 
 \hline
 Linear Product Flow & \begin{tabular}[c]{@{}c@{}} Can be employed to implement \\ a diagnosis algorithm from \\ perspective of product \end{tabular}  \\
 \hline
 Time Stamps on Data & \begin{tabular}[c]{@{}c@{}} Internal time stamps on sensor \\ data can be used to track  \\ correct processing of product \end{tabular} \\
 \hline
 Quality Control & \begin{tabular}[c]{@{}c@{}} Direct quality control \\ enables easier diagnosis \\ \end{tabular}  \\
 \hline
 \begin{tabular}[c]{@{}c@{}} Single \\ Automation Software \\\end{tabular} & \begin{tabular}[c]{@{}c@{}} Only one data stream \\ needs to be processed \\ \end{tabular}  \\ [1ex] 
 \hline
\end{tabular}
\end{center}
\end{table}

As summarized in Table 1, RIMs have several characteristics which can be exploited for diagnosis. As mentioned before, the product flow can be seen as linear. Therefore, for a diagnosis algorithm from the perspective of the product, the rotating motion of the machine can be ignored. A benefit of RIMs is that, since all tooling operations on the product are controlled by the same automation software, only one data stream needs to be processed and analysed and the time stamps of the values in this data stream are synchronized. Keeping track of time-sensitive steps is therefore easier than in set-ups where different machines work on the same product and their time stamps might not be perfectly synchronized. Furthermore, since RIMs often have even the quality control on the same machine, also these data points are synchronized with the other data points from production and can be easily mapped to the product path.

\section{Diagnosis Algorithm}

In the following, a diagnosis algorithm for a RIM will be developed, using the example of the machine modeled in the previous section. While the evaluation will be done using this specific machine, the approach can be generalized to other RIMs. 

\subsection{Formalization of the Problem}

First, the problem of diagnosis in RIM needs to be formalized. For this, the processes within the RIM need to be described formally. Since the algorithm will focus on the product, only the product states need to be taken into account. 

\begin{definition}[Product State]
\label{def:state}
The product state $S_i \in S$ with $S =\{1, \dots, n\} \subset \nat$ for $n \in \nat $ is the condition of the product at a given time  $t = t_s$. 
\end{definition}

A product's description is not only defined by the characteristics of the product at a defined time, but must also include its position in the RIM, since a product can have the same state $S_i$ while located at different positions within the RIM. 

\begin{definition}[Product Position]
\label{def:position}
The position $P_i$ of the product in the RIM is one element of a defined set of positions $P$, therefore $P_i \in P$ with  $P =\{1, \dots, n\} \subset \nat$ for $n \in \nat $. At each position $P_i$ a defined number of state transitions occur (see Definition 3). 
\end{definition}

With the \textit{Product State} and the \textit{Product Position}, a product can be fully described. To further describe the RIM, transitions between different states and positions need to be possible.

\begin{definition}[State Transition]
\label{def:transition}
A state transition $T_i \in T$ with  $T =\{1, \dots, n\} \subset \nat$ for $n \in \nat $ is a change in the products characteristics, leading to a change in the state of the product from state $S_i$ to state $S_j$. At each product position $P_i$, a defined number of state transitions happen. 
\end{definition}

For each state transition, a certain tool in the RIM is necessary. Each of these tools can have one ore more sensors attached to it, which report the position and other parameters of the tool. A state transition describes changes to the products characteristics, but not changes in position within the RIM. 

\begin{definition}[Rotation]
\label{def:rotation}
The rotation $R_i$ with $R_i \in R$ with  $R =\{1, \dots, n\} \subset \nat$ for $n \in \nat $ changes the product position from a position $P_i$ to the next position $P_j$ within $P$.  
\end{definition}

Using the Definitions 1 to 4, the path of the product through the machine and its final state can be fully described. 

The diagnosis algorithm should keep track of the path of the product through the machine and compare the expected state and position of the product with the reported values from the machine. This can be done step-wise, where each production step is the sum of all state transitions happening between two rotations. 

\begin{definition}[Production Step]
\label{def:step}
A production step $Y_i \in Y$ with  $Y =\{1, \dots, n\} \subset \nat$ for $n \in \nat $is the entirety of all state transitions $T_n, \dots, m$ with $n,m \in \nat $ taking place between two rotations $R_i$ and $R_j$. 
\end{definition}

A production step happens at a certain, defined position within the RIM, therefore the product position $P_i$ equals the production step $Y_i$. The relation between state $S$, state transition $T$, position $P$, production step $Y$ and rotation $R$ is shown in Figure 2.

\begin{figure*}[h!]
  \centering
  \includegraphics[width=.9\textwidth]{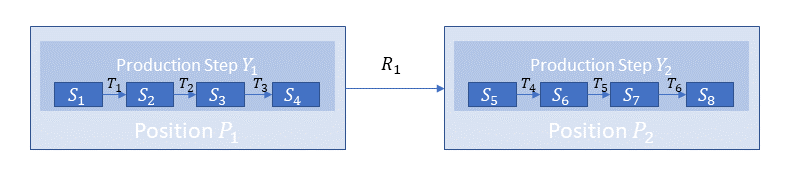}
  \caption{\label{Abb_Formal} Formalization of the RIM. At each position $P_i$ a defined number or state transitions $T_i$ happen. The sum of these state transitions $T_1, \dots, n$ is referred to as a production step $Y_i$. Once a production step is reported as finished, the rotation $R_i$ moves the product to the next position.}
\end{figure*}

Since RIMs carry out production steps in a clocked order, time is important for the analysis of the processes within the machine. However, in most cases the absolute time is not as important as the time between production steps. Therefore, the time a product spends in each production step is more important for the diagnosis than the absolute time. 

\begin{definition}[Internal Time]
\label{def:step}
The internal time $t_i$ is the time between the product entering the machine and the final product exiting the machine. The internal time $t_i$ starts with $t_i = 0$ when the product enters the first production step of the RIM. 
\end{definition}

Each product state is linked to a certain machine part, whose action triggered the state transition of the product at a given internal time. These machine parts report their actions via sensors in the RIM, including a time stamp. The state of a product is therefore not described directly, but can only be derived from the values of the sensors of the machine parts acting on the product. The actions which need to be taken by the machine to transform the product are pre-defined and their expected effects on the product are known. 

\begin{definition}[Process Description]
\label{def:process}
The process description $M$ is a tuple $(O,U,V,W)$, where
\begin{itemize}
    \item $O$ is a vector $(o_1,o_2,\dots,o_i) $ with $i,o \in \nat $ of the order of all production steps $Y_i \in Y$ within the RIM
    \item $U$ is a vector $(u_1,u_2,\dots,u_n) $ with $n \in \nat $ and $u \in \real $ of all timings within the RIM in internal time
    \item $V$ is a tuple of vectors $V = (v_1,v_2,\dots,v_k)$ with $k= (1,\dots,T_i)$, which is the mapping of all sensors onto a state transition $T_i$
    \item $W$ is a tuple of vectors $W = (w_1,w_2,\dots,w_k)$ with $k= (1,\dots,T_i)$, which is the mapping of all sensors measuring tool positions onto a tool within the RIM whose state they report
\end{itemize}
\end{definition}

The process description does not only define the order of steps within the production process, it also needs to contain the information about what the sensors measure and how this relates to the state transitions within the individual production steps. 
Within this process description, all sensors must be mapped onto the state transition in which they measure a machine parameter. Several sensors can be mapped onto one state transition, for example when one sensor measures the positions of a hydraulic cylinder and a second the pressure this cylinder exerts onto the product. The process description for this case should be step-wise, meaning each production step $Y_i$ (see Definition 4) should be described individually. 

\begin{definition}[Step-wise Diagnosis Problem]
\label{def:step}
The step-wise diagnosis problem is a tuple $(M, E, K)$, where
\begin{itemize}
    \item $M$ is the process description according to Definition \ref{def:process}
    \item $E$  is a vector $(e_1,e_2,\dots,e_n) $ with $e \in \real $ and $n \in \nat $ is the set of expected sensor values
    \item $K$ is a vector $(k_1,k_2,\dots,k_n) $ with $k \in \real $ and $n \in \nat $, reported by the sensors in the RIM, including a time stamp
\end{itemize}
\end{definition}

The task of step-wise diagnosis is to compare the set of expected values $E$ with the set of actual values $K$ for each production step and identify any inconsistencies between $K$ and $E$. Once an inconsistency is found, a possible cause of this inconsistency should be identified using the process description $M$.

Since the data in one production step is limited to sensors that report on actions which happened in this production step, it is possible that faults are detected that cannot be definitely explained, for example when multiple diagnoses are possible for an observed fault. In such cases it can be useful to use data from multiple production steps to identify the root cause of a fault. 

\begin{definition}[Multi-step Diagnosis Problem]
\label{def:step}
The multi-step diagnosis problem is a tuple $(M, E, K, Z)$, where
\begin{itemize}
    \item $M$ is the process description according to Definition \ref{def:process}
    \item $E$  is a vector $(e_1,e_2,\dots,e_n) $ with $e \in \real $ and $n \in \nat $ containing the expected sensor values
    \item $K$ is a vector $(k_1,k_2,\dots,k_n) $ with $k \in \real $ and $n \in \nat $, reported by the sensors in the RIM, including a time stamp
    \item $Z$ is a vector $(z_1,z_2,\dots,z_n) $ with $n \in \nat $, in which the possible diagnoses can be saved
\end{itemize}
\end{definition}

\subsection{Description of the Algorithm}

In the following, the implementation of two different algorithms is described. The first algorithm is a step-wise diagnosis algorithm, solving the diagnosis problem described in Definition 8, while the second algorithm is a multi-step diagnosis algorithm, which addresses the problem shown in Definition 9. 

Since the machine runs at a high speed and it is not necessary to diagnose the production of a product which passes through the machine without problems and is not later identified as problematic, the diagnosis algorithm is only activated once a fault is registered in a product. This is most often the case in one of the quality controls. However, a product can also successfully pass the quality controls but still show faults which are only recognized later. In this case, the diagnosis algorithm might be used at a later time-point to trace the path of the product through the machine. It would therefore be necessary to safe data from the RIM for later diagnosis.

In the case described here, the anomaly detection is on the level of the product, therefore an anomaly is detected when a faulty product is registered. An anomalous product triggers the diagnosis algorithm. The diagnosis on the level of the product than aims at identifying the production step and, ideally, the tool that was responsible for the fault in the finished product as a root cause.

Both algorithms take as input a list of expected sensor values, a list of actual sensor values measured in the RIM and the process description, which contains information about the production steps and their order, the timings of the processes in the RIM, the mapping of sensors onto a state transition and a mapping of sensors onto tools, according to Definition 7.

\begin{algorithm}
\caption{Step-wise Diagnosis Algorithm}\label{alg:cap}
\begin{algorithmic}
\Require $finalproduct = faulty$
\For{<All steps in RIM>}
    \State  $E \gets Expected Values$       
    
    \Comment{load expected values}
    \State  $M \gets Process Description$     
    
    \Comment{load process description}
    \State  $K \gets Measured Values$      
    
    \Comment{load measured values}
    \If {\textit{checkSAT(E,M,K)}}
    
    \Comment{Check if expected values match measured}    
    
        \Return OK
    \Else
     \If {$fault = time fault$}
     
     \Comment{check if fault in timing}   
     
     \Return {"Error in step " + $step\_number$ + ": Timing fault"}
     \EndIf
        \State  $K_f \gets nameconflictR$   
        
        \Comment{save name of conflict value}
        \State {\textit{check M for $K_f$}}
        
        \Comment{find sensor for conflict value}   
        
        \Return {"Error in step " + $step\_number$ + ": Fault found"\\
    \hspace{10mm} "Most likely cause: " + $name\_conflict$}        
    
        \Comment{report sensor mapping value for conflict}
    \EndIf
\EndFor
\end{algorithmic}
\end{algorithm}

Algorithm 1 one shows the pseudo-code of the implementation of the step-wise diagnosis algorithm. The algorithm uses an SMT solver to check the validity of the observed values $K$ from the machine with respect to the expected values $E$. Since the observed values from the real machine will never be exactly equivalent to the expected ones, a tolerance range for the expected values is given, which can be defined symmetrical around the expected value or can be bigger in one direction than the other. 

The algorithm then compares expected to real values for each production step of the RIM. When an inconsistency is encountered, the cause for this inconsistency needs to be identified. First, the algorithm checks whether the inconsistency was between the expected values for the timings in the production steps and the actual timings. If this is the case, a timing error and the state transition in which it occurred is reported. In case the fault is not a timing fault, the corresponding sensor which measured the inconsistent value needs to be identified. Once this is done, the tool which is described by this sensor can be identified and reported as the root cause for the faulty product. 

Algorithm 2 describes the multi-step diagnosis algorithm. The basic algorithm is the same as Algorithm 1, the main difference is that Algorithm 2 also takes into account failures which cannot be explained within one step. This is possible when a fault can have multiple causes and this conflict cannot be resolved within one production step. In this case, the algorithm saves the possible explanation in a variable outside of the main loop, therefore building a memory of former possible root causes. In later steps, this is than used it to check whether an explanation can be found in later steps. For this, again, an SMT solver is employed, which uses the expected values and the measured values from the current step, together with the process description and the list of faults and possible explanations from earlier steps to attempt finding a definite explanation for these earlier faults. 

The advantage in using an SMT solver for this approach is that the implementation of the algorithms only needs minimal changes when the machine changes, since the input into the algorithm can be implemented externally. The process description and lists of expected sensor values can be implemented outside of the main algorithm, so that the approach can be easily adapted. 

Both algorithms try to find the cause for the anomaly detected in the product. Therefore, the algorithms diagnose the process which led to the anomaly in the final product. However, the cause for the anomaly in the product is an anomaly in the production process. 

\begin{algorithm}
\caption{Multi-Step Diagnosis Algorithm}\label{alg:cap}
\begin{algorithmic}
\Require $finalproduct = faulty$
\State $Y \gets empty list $

\For{<All steps in RIM>}
    \State  $E \gets Expected Values$   
    
    \Comment{load expected values}
    \State  $M \gets Sensor Mapping$ 
    
    \Comment{load process description}
    \State  $K \gets Measured Values$    
    
    \Comment{load measured values}
    \State  $Z \gets empty$    
    
    \Comment{initialize vector for diagnoses}
    \If {\textit{checkSAT(E,M,K)}}
    
        \Return OK
    \Else
    \If {$fault = time fault$}
     
     \Comment{check if fault in timing}   
     
     \Return {"Error in step " + $step\_number$ + ": Timing fault"}
     \EndIf
        \State  $K_f \gets nameconflictR$    
        
        \Comment{save name of conflict value}
        
        \State {\textit{check M for $K_f$}}
        
        \State  $Counter \gets +1$

        \Comment{count number of possible explanations}
    \EndIf
        \If {\textit{$C == 1$}}
        
        \Return{"Fault found in step" + $step number$}
            {"Most likely cause" + $M_f$}
        
        \Comment{report sensor mapping value for conflict}
        \Else
        \State $Z \gets K_f$
        
        \Return {"Fault found in step" + $step number$}
        {"More than one explanation possible"}
        \EndIf
    \If {\textit{More than one explanation found}}

    \State  \textit{checkSAT(E,M,K,Z)}
    
    \Return {"Explanation for fault in earlier step found!"}
    \EndIf
\EndFor
\end{algorithmic}
\end{algorithm}

\section{Evaluation}

To evaluate the algorithms described in the preceding section, they were implemented in python. As a SMT solver the Z3 solver \cite{de2008z3} was used. The description of the RIM as provided by our cooperation partners was used to generate the process description, as well as the expected values for the sensor measurements and the expected times for each state transition. The model of the RIM described in Section 3.1 was used for simulations containing faults which the algorithms were supposed to diagnose. 

Figure 3 shows two examples for faults and how they can be diagnosed. In Figure 3A, the diagnosis algorithm is triggered when station 8 reports a product as Not-OK. This is due to station 6 reporting that the product failed the tightness test. The algorithms then check each station for consistency. When an inconsistency is found, Algorithm 1 reports it directly, while Algorithm 2 saves the possible explanations and checks whether another explanation is possible. In Figure 3A, only one explanation is possible and both algorithms would report it. In Figure 3B, however, two possible explanations for the same observation are possible. Also here, the product fails the tightness test, however in station 4, the measured values show that the pressure on the product was incorrect. This can be explained by a broken jack cylinder or by a wrong position of a product part. In this step, this cannot be answered and Algorithm 1 would report both explanations. Algorithm 2 would save the explanations until the next step in the diagnosis, where it can be determined that the product was in the wrong position and this is then reported as the diagnosis. 

As mentioned above, the diagnosis from the product perspective is triggered when a product shows an anomaly, mostly in the quality control. This then activates the diagnosis algorithm which aims at identifying the root cause of the anomaly observed in the product. The anomaly in the product is, however, caused by an anomaly in the machine behavior.

For evaluation of the two algorithms, five faults were simulated. Table 2 summarizes the results of the evaluation. Both algorithms performed well when the cause of the fault could be identified within a single step of the RIM. Algorithm2 could additionally identify the root cause of a fault whose cause could only be identified by combining the knowledge from two different steps. Both algorithms failed to identify when a sensor in the quality control was broken and when a broken piece was entered into the assembly process. 

\begin{table}[]
 \label{solutionstable}
 \caption{Faults and Diagnoses by Algorithms}
 \centering
\begin{center}
\begin{tabular}{||c | c | c||} 
 \hline
 Fault & Algorithm 1 & Algorithm 2 \\ [0.5ex] 
 \hline\hline
 Timing Jack Cylinder & \checkmark & \checkmark \\ [0.5ex] 
 \hline
 Part in Wrong Position & $\times$ & \checkmark  \\[0.5ex] 
 \hline
 Pressure Sensor Broken & $\times$ & $\times$ \\[0.5ex] 
 \hline
 Jack Cylinder Broken & \checkmark & \checkmark  \\[0.5ex] 
 \hline
 Part Broken & $\times$ & $\times$  \\[0.5ex] 
 \hline
\end{tabular}
\end{center}
\end{table}

\section{Discussion}

In this study, we addressed the topic of diagnosis of RIMs. Analyzing the characteristics of RIMs, we found that they offer some features which can be exploited for the development of diagnosis algorithms specifically suited for diagnosing causes of faults in products processed on these machines. Based on the identified characteristics, we developed two diagnosis algorithms. These are able to identify faults in the RIM which affect the product quality. As was shown in Table 2, both algorithms are able to identify faults whose root cause can be definitely identified within one production step. This does not mean that the fault needs to be detected in the same step. Faults are often only reported once the product is sorted as Not-OK. Once a product is reported as Not-OK, the algorithm start moving through the production process backwards. The reason for the sorting into Not-OK is usually found in the quality control stations, however these do only report which characteristics of the product did not fit with the expectations, but do not identify a reason for these faults. To identify the root cause, the algorithms check the expected sensor values for each station and compare them to the actual measurements. They also take into account the expected and actual times for each state transition and rotation. When Algorithm 1 encounters an inconsistency, it will check the system description for the sensor that is mapped on the inconsistent value and report it as cause. Should two possible explanations be possible in one step, Algorithm 1 cannot determine which explanation is more likely and will report both. In such cases, Algorithm 2 has an advantage, since it can save this information and check whether information from other (earlier) stations can identify the actual root cause. However, this is only possible in one direction, since the algorithm moves backwards through the production process. Therefore, only earlier production steps can be used to add further information for identification of a root cause. 
As can be seen in Table 2, both Algorithms fail for certain types of faults. There are two main reasons for this. The first is the absence of certain sensors. No sensors are installed to measure vibrations or to check whether parts are damaged before entering the machine. In the real RIM, a system exists that checks one part of the final product before putting into the assembly station. However, this was not part of this model. Other parts are also not checked for damages. Therefore, these types of faults cannot be detected. The machine will perform the assembly steps on them and in the quality control, they will show faults. But, since no sensors report the damage, the algorithms cannot identify this as root cause. Another type of fault that cannot be identified is a broken sensor. In the case tested here, a sensor in the quality control was broken and reported a product as Not-OK, despite the product being OK. This could also not be diagnosed by the algorithms. 
Since algorithm 2 performed better and is able to identify more complex faults, it would be most useful to continue development on this algorithm for usage in the actual RIM. 

As pointed out before, the diagnosis to identify the root cause of an anomalous product needs to identify which anomaly in the production process lead to the anomaly in the product. Therefore, both algorithms were implemented to identify anomalies within the production process which are in turn the root cause for the anomaly in the product. 

\begin{figure*}[h!]
  \centering
  \includegraphics[width=1.0\textwidth]{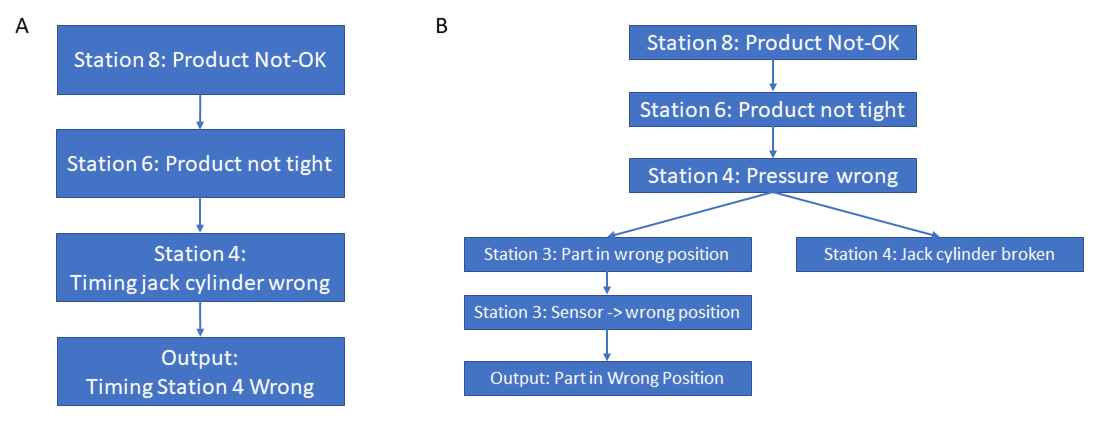}
  \caption{\label{Abb_Diag} Two examples of faults diagnosed by the algorithm. Figure A shows an example of a fault that can be diagnosed with only one possible explanation. At first, Station 8 reports that a product was Not-OK, this can be traced back to the quality control in station 6 measuring that the product was not tight. This defect can be traced to station 4, where a timing error in the jack cylinder occurred. Since no other explanation is found, this explanation is reported. Figure B shows an example of a failure with two possible explanations. Also here, the product is reported as Not-OK, which is explained by the measurement in station 6. Station 4 reports that the pressure was wrong during the assembly process. For this, two possible explanations are possible in the machine. The jack cylinder could be broken or the part could have been in the wrong position from the preceding station. While Algorithm 1 would report both diagnoses, Algorithm 2 would use the observations from station 3 to determine that the part was in the wrong position, while the jack cylinder is not broken. In Algorithm two only the diagnosis that the part was in the wrong position would be reported.}
\end{figure*}

\section{Conclusion and Outlook}

Despite their widespread application, almost no research exists on the diagnosis of RIMs, especially considering the product perspective. A possible explanation for this is the fact that RIMs are often built and employed by small and medium seized companies who lack the possibilities to develop anomaly detection or diagnosis algorithm on their own. Nevertheless, RIMs are ideally suited for diagnosis algorithms, since they incorporate processing and quality control of products into one machine, which means that all data will be generated within the same automation unit and can be reported in one file with one time stamp. This, and the fact that the product flow can be simplified into a linear product flow, makes diagnosis of possible faults in the products simple (\textit{RQ1}). Since SMT solvers have been established for the diagnosis of various processes, including production processes, their use seems plausible also for the case of RIMs. However, to the best of the author's knowledge, such algorithms haven't yet been developed for RIMs. Here, two diagnosis algorithms based on SMT solvers for RIMs have been suggested (\textit{RQ2}). The evaluation showed that both algorithms can diagnose faults whose root cause can be found in one production step, while Algorithm 2 is also able to diagnose faults whose root cause identification needs observations from several production steps. Some faults cannot be diagnosed with these algorithms, mainly faults that occur when a sensor is broken or when a damaged piece is placed into the machine. In general, the diagnosis algorithms are heavily limited by the available sensors. It is, for example, not possible to identify faults due to vibrations, since no vibration sensors are installed. 
In this work, the algorithms were only tested on a model of the RIM. In the future, this algorithm should be tested on a real machine. Furthermore, more sensors should be integrated into the algorithm and dependencies between new sensors should be added to the system description. Further research should also implement a system description for a wider range of RIMs to prove the applicability of the algorithms to RIMs in general. Another feature to be added is the ability of the algorithm to use information not only in one direction.

\newpage
\section*{Acknowledgments}

This work was funded by the "Zentrales Inoovationsprogramm Mittelstand" of the Bundesministerium für Wirtschaft und Energie within the project "RepAssist: Automatische Diagnose- und Reparaturassistenz für mehrstufige Produktionsanlagen anhand von Anlagenkausalitäten"

\newpage

\newpage
\bibliographystyle{unsrt}
\bibliography{dx23}

\end{document}